\documentclass[runningheads]{llncs}

\usepackage[T1]{fontenc}
\usepackage[utf8]{inputenc}
\usepackage{graphicx}
\usepackage{amsmath,amssymb}
\usepackage{booktabs}
\usepackage{url}
\usepackage{microtype}
\usepackage{etoolbox}
\usepackage[hidelinks]{hyperref}

\AtBeginDocument{\renewcommand{\doi}[1]{%
  \href{https://doi.org/\detokenize{#1}}{\nolinkurl{doi:#1}}}}

\makeatletter
\patchcmd{\@maketitle}{\vskip .8cm\if!\@subtitle!}{\vskip .45cm\if!\@subtitle!}{}{}
\patchcmd{\@maketitle}{\@author\vskip.35cm}{\@author\vskip.15cm}{}{}
\makeatother

\begin{document}

\title{Benchmarking Generative Trajectory Models for Active-Inference Control}
\titlerunning{Generative Trajectory Models for AIF Control}
\author{Yulin Li\inst{1}\orcidID{0009-0001-0135-4332} \and
Mohsen A. Jafari\inst{1}\orcidID{0000-0003-4776-0632} \and
Andrea Matta\inst{2}\orcidID{0000-0003-3902-2007}}
\authorrunning{Y. Li et al.}
\institute{Rutgers University--New Brunswick, Piscataway NJ 08854, USA \and
Politecnico di Milano, Via La Masa 1, 20156 Milano, Italy}

\maketitle
\begingroup
\renewcommand\thefootnote{}
\footnotetext{Code: \url{https://github.com/lyeeonardo/generative-trajectory-benchmark}}
\addtocounter{footnote}{-1}
\endgroup

\vspace{-6pt}
\begin{abstract}
Learning from trajectory demonstrations offers a route to active-inference control of complex systems whose dynamics are difficult to model explicitly. We introduce generative active-inference control (GenAIF), in which one generative trajectory model learns from demonstrations and measured action interventions to supply a goal-conditioned policy distribution and a state-to-observation likelihood mapping. From this control design, we derive three model requirements: (i) useful action proposals, (ii) accurate prediction under imposed actions, and (iii) probabilistic observation evidence for belief updating and expected information gain. We benchmark diffusion, autoregressive Transformers, conditional variational autoencoders (CVAEs), and flow matching in a MuJoCo manipulation task with multiple physical conditions. Diffusion delivers the strongest control across the tested dynamics, while CVAE combines comparable short-horizon prediction with much faster inference. Correct conditioning is decisive, and trajectory reuse offers further computational savings. In replay after an unannounced tilt change, pretrained diffusion updates belief fastest among the original models; fine-tuning on recovery demonstrations further accelerates identification and sustains accurate tracking. These findings support the use of shared generative trajectory models to connect action proposal, controlled prediction, and observation evidence within GenAIF.

\end{abstract}

\section{Introduction}
\label{sec:introduction}

People learn how to act by trying, observing, and adjusting. Watching someone learn a movement can help us adapt to the same unfamiliar physical dynamics \cite{Mattar2005Observing}. Active Inference (AIF) offers an elegant account of how organisms bring perception and action together: an internal generative model connects hidden causes to sensory observations, and perception updates beliefs by minimizing variational free energy (VFE), an upper bound on sensory surprise \cite{Friston2010FreeEnergy,Friston2017ProcessTheory}. Action selection looks ahead through expected free energy (EFE), combining preferences for useful, safe outcomes with the information that an action is expected to reveal \cite{Parr2022ActiveInference,Friston2015Epistemic}.

Our goal is a general framework that learns how a complex system behaves and how to control it from trajectory demonstrations. It should accommodate prior knowledge of system structure and hidden dynamics when available, and eventually learn their structure, form, and range when such knowledge is limited or absent.

Realizing this goal is computationally demanding: the controller must predict uncertain consequences while searching continuous action sequences \cite{Tschantz2020Scaling,Fountas2020DeepAIF}. Learned proposals can concentrate a finite EFE-evaluation budget on useful futures. We first explored this route in autonomous parking, combining diffusion-generated strategies with AIF evaluation and observation-driven adjustment \cite{Huang2026ProbDiffusion}. Although the approach generalized to new parking settings, its design was tailored to that task. This experience led us to ask how much of the active-inference loop a single learned trajectory model could support.

\emph{Can a single learned generative trajectory model provide both a proposal distribution over policies and action-conditioned predictions of future observations sufficient for belief updating and expected-free-energy evaluation in active inference?}

We propose generative active-inference control (GenAIF) around this shared model of action--observation trajectories. The same learned parameters support candidate generation and prediction under supplied actions; predictive evidence then supports belief updates about hidden operating conditions. The controller uses these operations to alternate between observing, updating its belief, and selecting actions according to task preferences and expected information gain. We learn the model offline from demonstrations supplemented with measured action interventions. At this stage, the hidden variables and their possible values are specified.

This design leaves a concrete model-selection problem. Which capabilities must a trajectory generator provide, and which generative families can provide them together at a useful computational cost? We identify three requirements: actionable trajectory proposals, accurate prediction of observations under externally imposed actions, and predictive evidence that can distinguish hidden conditions using actual observations. A plausible generated trajectory does not by itself establish the latter two capabilities. These requirements give us a way to test our framework's learned component and offer other researchers practical criteria for combining AIF with generative trajectory models.

We design two groups of experiments in a custom MuJoCo ball-pushing environment to evaluate diffusion, autoregressive Transformers, conditional VAEs, and flow matching. The first benchmarks the required model capabilities and computational cost across different physical conditions. The second tests belief tracking after a hidden dynamics change using replayed actions and observations, comparing the four pretrained models and a diffusion variant fine-tuned on recovery demonstrations. At the tested settings, diffusion gives the strongest direct control, while CVAE provides similar short-horizon position prediction at lower cost. Reuse reduces computation with model-dependent effects on control. Recovery fine-tuning strengthens diffusion's observation evidence, enabling faster and more stable inference of the changed dynamics.

The contributions of this paper are as follows: (1) we propose GenAIF, in which a single model learned from trajectory demonstrations supplies a goal-conditioned policy distribution and a state-to-observation likelihood mapping; (2) we develop a MuJoCo environment and benchmark to evaluate the capabilities required of this shared model; and (3) we compare four generative model families in performance and computational cost, and demonstrate belief updating under changing hidden dynamics through replay, including the benefit of recovery fine-tuning.


\section{GenAIF Framework}
\label{sec:framework}

\smallskip
\noindent\textbf{Learning a shared generative model.}

Let \(s_t\) contain the physical variables and operating conditions needed to predict the system's evolution. Some coordinates are observed, others hidden. The controller applies action \(a_t\) and receives observation \(o_{t+1}\). We learn a conditional generative trajectory model offline from labeled demonstrations and measured action interventions. The learned parameters \(\theta\) remain fixed during control.

For horizon \(H\), write future actions as \(U=a_{t:t+H-1}\), states as \(S_+=s_{t+1:t+H}\), and observations as \(Y=o_{t+1:t+H}\). We represent candidate policies by action sequences \(U\), replanned after each observation. With proposal context \(c_k=(s_k,h_k,g)\), a causal factorization of the shared model is
\begin{equation}
p_\theta(U,S_+,Y\mid c_t)=\prod_{k=t}^{t+H-1}
\mu_\theta(a_k\mid c_k)\,
T_\theta(s_{k+1}\mid s_k,a_k)\,
O_\theta(o_{k+1}\mid s_{k+1},a_k).
\label{eq:framework_joint}
\end{equation}
Here \(h_t=(o_0,a_0,\ldots,a_{t-1},o_t)\) is the available history, extended by generated actions and observations at future steps. The factors describe action generation (\(\mu_\theta\)), state transitions (\(T_\theta\)), and observations (\(O_\theta\)); they can share learned parameters. The task specification \(g\) guides proposals and preferences, while physical predictions depend on the system and imposed actions.

\begin{samepage}
For interacting subsystems, let \(s_t=(s_t^1,\ldots,s_t^m)\). A known, fixed graph \(\mathcal G\) specifies the parents of each next-state component. Under conditional independence given these parents and the action,
\begin{equation}
T_\theta(s'\mid s,a,\mathcal G)
=\prod_{i=1}^{m}T_{\theta,i}\!\left(s'^i\mid s^{\mathrm{Pa}_{\mathcal G}(i)},a\right).
\label{eq:framework_graph}
\end{equation}
Parents can include the component itself. This allows hidden conditions to persist, drift, or switch within a fixed transition model, provided those transitions are represented in data or prior knowledge. Graph structure is optional; we suppress \(\mathcal G\) in the remaining notation.
\end{samepage}

The shared model must therefore support two queries: predict the next state from the current state and an imposed action, and generate an \(H\)-step trajectory from the current state, history, and task condition \(g\):
\begin{equation}
\begin{aligned}
s_{t+1}&\sim T_\theta(\cdot\mid s_t,a_t),\\
(U,S_+,Y)&\sim p_\theta(\cdot\mid s_t,h_t,g).
\end{aligned}
\label{eq:framework_queries}
\end{equation}
For this benchmark, measured physical variables are fully observed, and the representation and range of hidden dynamics are known. The first experiment group supplies tilt; the belief-tracking experiment infers lateral tilt among three known values using a prescribed context-transition matrix. The following equations describe the general inference-and-control loop; the benchmark specializes this formulation to inference over hidden tilt.

Within each model family, proposal and prediction reuse the same trained trajectory backbone or decoder, operating on alternating action and observation tokens. Queries differ in role, known-coordinate mask, and conditioning. Proposal queries generate future actions and observations using task and quality conditions. Prediction queries fix the supplied actions, remove goal and quality conditioning, and generate observations under the physical context. Calibrated predictive samples supply approximate observation likelihoods without a separately trained dynamics model. In AIF terms, these operations provide a goal-conditioned policy distribution and an approximate state-to-observation likelihood mapping, conditioned on history and imposed actions.

\smallskip
\noindent\textbf{Inferring the current state.}

The controller maintains a belief \(b_t(s)\approx p_\theta(s_t=s\mid h_t)\) over the current state \cite{DaCosta2020DiscreteAIF}. After executing \(a_t\), it predicts how that state evolves and corrects the prediction using the actual observation:
\begin{equation}
\begin{aligned}
b^-_{t+1}(s')&=\int T_\theta(s'\mid s,a_t)b_t(s)\,ds,\\
b_{t+1}(s')&\propto O_\theta(o_{t+1}\mid s',a_t)b^-_{t+1}(s').
\end{aligned}
\label{eq:framework_filter}
\end{equation}
This correction minimizes variational free energy over normalized beliefs \(q\):
\begin{equation}
F_{t+1}[q]
=D_{\mathrm{KL}}(q\|b^-_{t+1})
-\mathbb E_q[\log O_\theta(o_{t+1}\mid s',a_t)].
\label{eq:framework_vfe}
\end{equation}
The first term retains information from the predicted belief; the second rewards explaining the observation \cite{Friston2017ProcessTheory}. An unrestricted \(q\) recovers the Bayesian posterior; a restricted family approximates it. Online adaptation therefore changes the state belief while keeping the learned model fixed.

\smallskip
\noindent\textbf{Proposing and selecting actions.}

Given \(b_t\), we sample state hypotheses \(s^{(j)}\sim b_t\), generate joint trajectories using Eq.~\eqref{eq:framework_joint}, and retain their action sequences \(U^{(j)}\). To evaluate a candidate, we impose its actions on the model and predict across the same full belief:
\begin{equation}
Q_t^U(\mathbf S,Y)
=b_t(s_t)\prod_{k=t}^{t+H-1}
T_\theta(s_{k+1}\mid s_k,a_k)\,
O_\theta(o_{k+1}\mid s_{k+1},a_k),
\label{eq:framework_prediction}
\end{equation}
where \(\mathbf S=s_{t:t+H}\). Removing the action-generation factors \(\mu_\theta\) represents the intervention \(\operatorname{do}(U)\). Every candidate is evaluated under the common belief, including hypotheses other than the one that generated it.

We use the following cost-augmented approximation to EFE for action selection \cite{Parr2022ActiveInference,Friston2015Epistemic}:
\begin{equation}
G_t(U)=
\underbrace{\mathbb E_{Q_t^U}[C(U,Y;g)]}_{\text{expected task cost}}
-\lambda\underbrace{I_{Q_t^U}(\Xi;Y)}_{\text{expected information gain}},
\qquad \lambda\geq0.
\label{eq:framework_score}
\end{equation}
Here \(C\) specifies task and action costs, \(\Xi=\chi(\mathbf S)\) is the hidden state quantity of interest, and \(\lambda\) sets the value of information in cost units. Mutual information measures how much future observations are expected to reduce uncertainty about \(\Xi\). Its predictive likelihood must also support the update from actual observations.

Among candidates satisfying action bounds and a predicted-risk limit, select the one with minimum \(G_t\), execute its first action, observe, and update the belief. This closes the loop: the belief shapes proposals and their evaluation, while the selected action produces the evidence for the next belief. All hypothetical outcomes come from the learned model.

Shared masked queries and calibrated predictive likelihoods approximate the framework's distributions in the benchmark. The belief-tracking experiment evaluates the original and recovery-fine-tuned models on replayed actions and observations.

\section{Required Capabilities and Candidate Models}
\label{sec:capabilities_models}

Our finite-candidate GenAIF controller imposes three evaluation requirements. The two queries in Eq.~\eqref{eq:framework_queries} supply candidates and their predicted consequences, but using predictions for belief updating and expected information gain also requires reliable observation evidence. We therefore assess proposal usefulness, controlled prediction, and evidence quality separately; the latter two are different demands on the same predictive model.

\smallskip
\noindent\textbf{Action proposals.}
With a finite action-search budget, the proposal distribution determines which action sequences are evaluated using \(G\). Even perfect scoring cannot select an action that was never proposed. We therefore require useful, executable action sequences.

\smallskip
\noindent\textbf{Controlled prediction.}
A proposal generated assuming normal lateral tilt (\(0^{\circ}\)) must also be evaluated under other lateral-tilt hypotheses in the current belief. This requires predicting the consequences of the same imposed actions across that belief, as in Eq.~\eqref{eq:framework_prediction}. Observations sampled alongside the original proposal represent only one possible future. Controlled predictions supply the outcome distribution needed to evaluate task cost, safety, and possible future observations.

\smallskip
\noindent\textbf{Observation evidence.}
A forecast must also say how strongly an observed outcome supports each hypothesis. Predictors with identical means can produce very different belief updates if their uncertainty differs. The predictive law must provide normalized likelihoods for the VFE update after execution and expected information gain in \(G\) before execution \cite{DaCosta2020DiscreteAIF}. Forecast accuracy and evidence quality therefore need separate evaluation. If hypotheses predict indistinguishable observations, the observation should not change their relative belief.

These requirements motivate four candidate families whose literature supports relevant proposal, prediction, or density-evaluation operations.

\smallskip
\noindent\textbf{Diffusion.}
Diffusion models~\cite{Ho2020DDPM,Peebles2023DiT} offer precedents for trajectory planning in Diffuser and Decision Diffuser, and action generation in Diffusion Policy~\cite{Janner2022Diffuser,Ajay2023DecisionDiffuser,Chi2023DiffusionPolicy}. CSDI demonstrates completion from observed time-series coordinates \cite{Tashiro2021CSDI}, motivating conditional prediction; score-based formulations also offer a route to likelihood evaluation \cite{Song2021ScoreBased}.

\smallskip
\noindent\textbf{Autoregressive Transformer.}
Transformers model sequential dependencies \cite{Vaswani2017Attention}. Trajectory Transformer demonstrates joint state--action modeling and planning \cite{Janner2021TrajectoryTransformer}, while Decision Transformer supports return-conditioned action generation \cite{Chen2021DecisionTransformer}. With probabilistic outputs, an autoregressive factorization offers both forecasts conditioned on supplied actions and evaluable conditional densities, motivating prediction and observation evidence alongside proposals.

\smallskip
\noindent\textbf{Conditional variational autoencoder.}
Conditional VAEs learn context-dependent distributions through latent variables \cite{Kingma2014VAE,Sohn2015CVAE}. Play-LMP provides evidence for goal-directed action generation \cite{Lynch2020PlayLMP}, and Trajectron++ for probabilistic forecasts conditioned on an ego-agent's planned motion \cite{Salzmann2020Trajectron}. These support the proposal and prediction hypotheses; using such predictions as observation evidence additionally requires accounting for latent-variable uncertainty.

\smallskip
\noindent\textbf{Flow matching.}
Flow matching provides conditional sampling and, through its continuous-flow formulation, a route to density evaluation \cite{Lipman2023FlowMatching}. Trajectory Conditional Flow Matching demonstrates forecasting and goal-conditioned planning \cite{Ye2024TrajectoryFlow}, while \(\pi_0\) generates robot action sequences \cite{Black2025Pi0}. These precedents motivate both prediction and proposal use.

The cited studies motivate the candidates without establishing all three capabilities in one checkpoint. Conditional completion alone does not establish accurate prediction under imposed actions or reliable observation evidence. Our benchmark tests shared proposal and prediction interfaces, then evaluates their evidence for inference, alongside supplied-condition sensitivity, trajectory reuse, and computation.

\section{Experimental Setup}
\label{sec:benchmark}

\begin{figure}[!t]
    \centering
    \includegraphics[width=\linewidth]{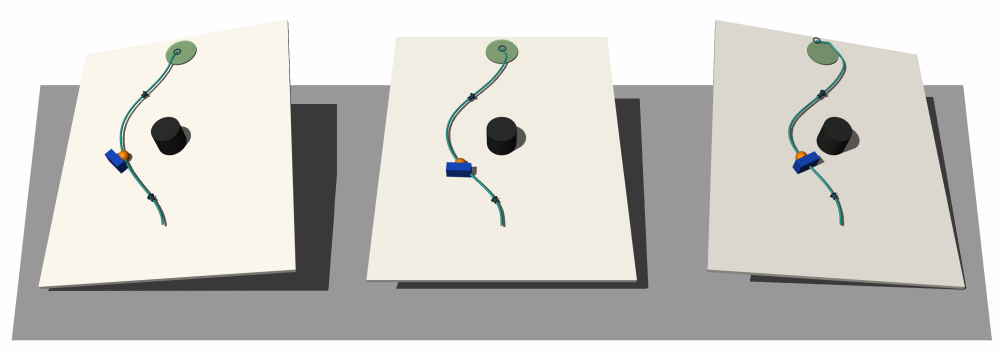}
    \caption{Ball-pushing task at illustrative lateral tilts. The pusher moves the ball around the obstacle toward the green goal; lines illustrate routes.}
    \label{fig:benchmark_environment}
\end{figure}

\noindent\textbf{Task.}
Our custom MuJoCo environment \cite{Todorov2012MuJoCo} requires a rectangular pusher to move a ball uphill, around an obstacle, and settle it safely in a goal region (Fig.~\ref{fig:benchmark_environment}). We vary lateral tilt across \(\{-15,0,+15\}^{\circ}\) and use three obstacle positions, giving nine physical conditions. The normal lateral tilt is \(0^{\circ}\); longitudinal tilt stays at \(+20^{\circ}\) in every condition. This tests whether one model can adjust its control to different dynamics while solving the same task. An episode ends on success, ball--obstacle collision, falling off the board, or the control-step limit.

Actions specify planar pusher velocity and yaw rate. The observation records ball position and velocity, pusher position, and pusher yaw. Each query receives five observations and four preceding actions, together with geometry, supplied tilt, and elapsed time; prediction horizons are \(H\in\{1,3,6\}\) steps. The simulator is deterministic, so repeated trials measure variation in generative sampling.

\smallskip
\noindent\textbf{Dataset.}
A feedback controller supplies expert demonstrations in every physical condition. From 216 successful reference episodes, we collect 65,195 short action interventions by restoring recorded simulator states and executing reference variants, perturbations, and exploratory actions. Demonstrations supply useful behaviour, while these interventions reveal what happens when actions depart from it. Prediction training retains unsafe outcomes; proposal training uses references and safe intentional continuations. Windows are truncated at physical termination, and padding is excluded from the loss.

Training, development, calibration, and test sets are separated by source rollout: each intervention stays with its reference episode. We remove exact duplicates and compute normalization from training data only. All sets cover the same physical conditions, so evaluation asks whether the models can handle new rollouts within this task. Private simulator state is never an input.

\smallskip
\noindent\textbf{Training and selection.}
All models complete the same training-data exposure budget. We select checkpoints on development performance in proposal usefulness, controlled prediction, and observation likelihood. Each selected checkpoint serves both proposal and prediction queries and remains fixed during evaluation. Test data influence neither selection nor calibration.

\begin{figure}[!t]
    \centering
    \setlength{\abovecaptionskip}{4pt}
    \includegraphics[width=\linewidth]{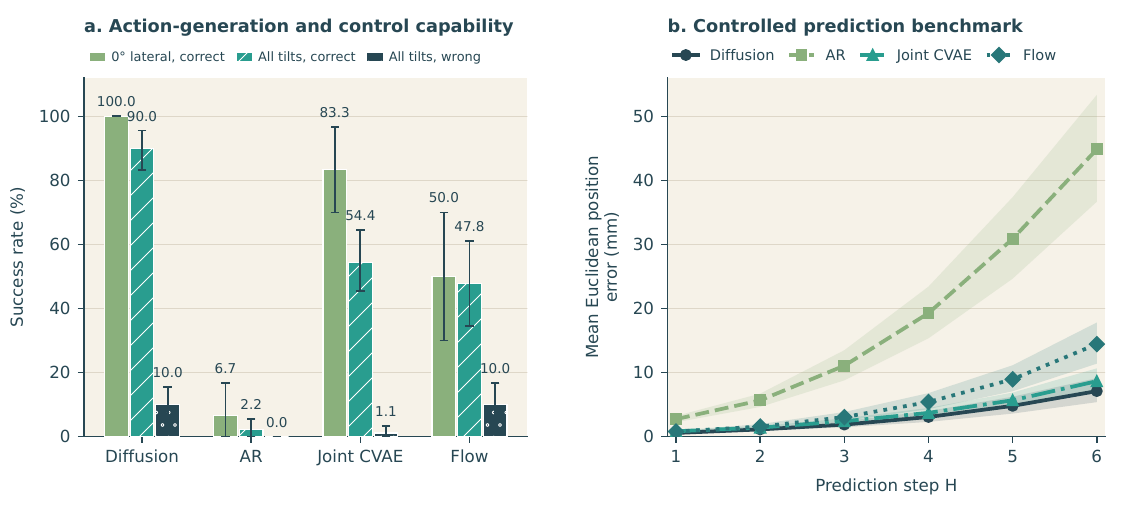}
    \caption{E1 control and prediction. (a) Control success with correct or incorrect tilt inputs (zero-lateral-tilt subset: 30 episodes; all tilts: 90 per group). (b) Mean Euclidean ball-position error of the predictive mean over six steps. Whiskers and shading show 95\% bootstrap intervals over 10 sampling seeds (a) and 18 source trajectories (b); bands are pointwise. AR: autoregressive Transformer.}
    \label{fig:benchmark_results}
\end{figure}

All models share the same training-data exposure budget. We retain the smallest tested model when additional capacity no longer improves development performance, since larger configurations reduced performance for some models. Diffusion continued to benefit from scaling, motivating its larger selected configuration. Both diffusion and flow matching are larger than CVAE and the autoregressive Transformer. Because architecture affects how complex trajectory information is represented, matching parameter counts alone would not ensure a fair comparison.

For belief updating, we construct Gaussian observation likelihoods from each model's predictive samples, treat yaw as circular, and calibrate uncertainty on a separate data split. This supplies a common way to turn forecasts into evidence about hidden conditions. We measure model evaluation time with synchronization on an NVIDIA RTX 5090. Detailed architectures, optimization settings, and sampling configurations are provided in the code.

\section{Experiments and Results}
\label{sec:results}

We evaluate the shared model in two experiment groups: the capability benchmark (E1) and the belief-tracking experiment (E2). In E1, direct execution tests proposal usefulness, imposed-action queries test prediction, and single-step condition identification tests observation evidence. We also evaluate condition sensitivity, trajectory reuse, and computation. E2 tests sequential belief inference after a hidden tilt change using replayed actions and observations, comparing the original frozen checkpoints and recovery-fine-tuned diffusion.

\begin{table}[!t]
    \caption{E1 control, prediction, and computation. Position RMSE uses 841 valid endpoints. Times cover model evaluation and checking per executed action, excluding simulator execution.}
    \label{tab:e1_results}
    \centering
    \small
    \setlength{\tabcolsep}{4pt}
    \begin{tabular}{lrrrrr}
        \toprule
        & \multicolumn{2}{c}{Success /90} & \(H=6\) position & \multicolumn{2}{c}{Time/action (ms)} \\
        \cmidrule(lr){2-3}\cmidrule(lr){5-6}
        Model & Direct & Reuse & RMSE (mm) & Direct & Reuse \\
        \midrule
        Diffusion & 81 & 68 & 14.295 & 57.123 & 9.794 \\
        Autoregressive & 2 & 2 & 65.809 & 48.735 & 12.982 \\
        CVAE & 49 & 61 & 15.264 & 5.643 & 1.236 \\
        Flow matching & 43 & 34 & 24.776 & 199.571 & 37.037 \\
        \bottomrule
    \end{tabular}
\end{table}

\smallskip
\noindent\textbf{Action generation across dynamics.}
We first test whether one learned model can control all nine physical conditions, using ten sampling seeds per condition for 90 episodes per setting. At each decision, it proposes one six-step trajectory, executes the first action, and replans. Tilt stays fixed within an episode. We compare correct and incorrect tilt inputs on matched scenes and seeds. This arm uses no online belief update or EFE-based candidate selection, so its success rate measures proposal usefulness under supplied conditions.

Diffusion provides the most reliable control across the tested dynamics: 90\% success, compared with 54\% for CVAE and 48\% for flow (Fig.~\ref{fig:benchmark_results}a). CVAE performs well in the zero-lateral-tilt cases but loses reliability across the broader set. Since the true tilt is supplied, these differences persist even without condition-estimation error. Accurate beliefs therefore need to be paired with a model that can turn the inferred condition into effective actions.

Conditioning is decisive: incorrect tilt inputs reduce diffusion's success by 80 percentage points, with large losses also seen for CVAE and flow. A capable generator can therefore choose ineffective actions when conditioned on the wrong dynamics, making accurate hidden-condition inference essential to the eventual controller.

The autoregressive model succeeds in only 2/90 episodes and remains unreliable even in the zero-lateral-tilt subset. This may reflect interference between dynamics during joint training and a loss of previously learned behaviour.

\smallskip
\noindent\textbf{Controlled prediction.}
We impose the same action sequences at sampled test states and compare predicted ball motion with simulator outcomes. Queries retain physical context but omit the desired goal and quality conditioning. This asks whether the model predicts what supplied actions actually do, independently of the actions it would propose. Figure~\ref{fig:benchmark_results}b follows position error across the prediction horizon.

Diffusion and CVAE produce the closest forecasts in both mean Euclidean position error (Fig.~\ref{fig:benchmark_results}b) and endpoint RMSE (Table~\ref{tab:e1_results}). At the sixth prediction step, mean Euclidean position error is 7.1\,mm for diffusion and 8.7\,mm for CVAE, compared with 14.4\,mm for flow and 44.9\,mm for the autoregressive model. Autoregressive error accumulates most rapidly, limiting extended action evaluation. CVAE's accuracy remains close to diffusion's despite weaker control: predicting consequences and generating useful actions are separate achievements.

\smallskip
\noindent\textbf{Computation and trajectory reuse.}
We compare generating a new trajectory at every decision with continuing the current trajectory while its predicted observations agree with actual motion. Both settings use the same physical cases and sampling seeds. Table~\ref{tab:e1_results} brings control success and endpoint prediction error together with model-plus-checker time per executed action. Timings include proposal re-prediction and exclude simulator execution.

CVAE has the lowest model-plus-checker time per executed action: 5.6\,ms, compared with 57.1\,ms for diffusion. Its accurate short-horizon predictions leave computational headroom for higher-frequency control. The choice is therefore between CVAE's speed and diffusion's stronger control across varying dynamics. Flow's tested solver makes it the most expensive option.

Reuse reduces model-plus-checker time per action by 73--83\%, but its value depends on the trajectory being followed. CVAE's success rises by 13 percentage points, while diffusion's falls by 14. Prediction agreement is a useful trigger for saving computation, but cannot by itself determine how long a plan should be followed. Reuse improves both speed and success for CVAE, while diffusion achieves higher control success with frequent regeneration. This suggests choosing the execution schedule jointly with the proposal model.

\smallskip
\noindent\textbf{Evidence for hidden-condition inference.}
Starting from equal belief in the three tilt hypotheses, each model predicts the same imposed action under each hypothesis and uses the observed transition to update their probabilities. This tests whether forecasts supply useful evidence about hidden dynamics, beyond the accuracy of their means. Table~\ref{tab:e1_evidence} reports identification accuracy, log loss of the true-tilt probability, and calibration error comparing confidence with accuracy across ten bins.

\begin{table}[!t]
    \caption{E1 hidden-tilt identification over 864 action--observation pairs. Lower log loss and calibration error indicate better probability estimates.}
    \label{tab:e1_evidence}
    \centering
    \small
    \setlength{\tabcolsep}{6pt}
    \begin{tabular}{lrrr}
        \toprule
        Model & Accuracy (\%) & Log loss (nats) & Calibration error \\
        \midrule
        Diffusion & 98.96 & 0.540 & 0.062 \\
        Autoregressive & 79.28 & 0.959 & 0.127 \\
        CVAE & 99.07 & 0.434 & 0.057 \\
        Flow matching & 99.07 & 0.474 & 0.069 \\
        \bottomrule
    \end{tabular}
\end{table}

Across 864 action--observation pairs, diffusion, CVAE, and flow achieve approximately 99\% condition-identification accuracy, compared with 79\% for the autoregressive model. CVAE also records the lowest log loss and calibration error, showing that probability quality differs even when identification accuracy is similar. These predictions provide the observation evidence required by GenAIF. E2 tests whether this evidence remains reliable across successive observations after a dynamics change, and whether recovery demonstrations improve belief inference when the resulting motion is unfamiliar.

\begin{figure}[!t]
    \centering
    \includegraphics[width=\linewidth]{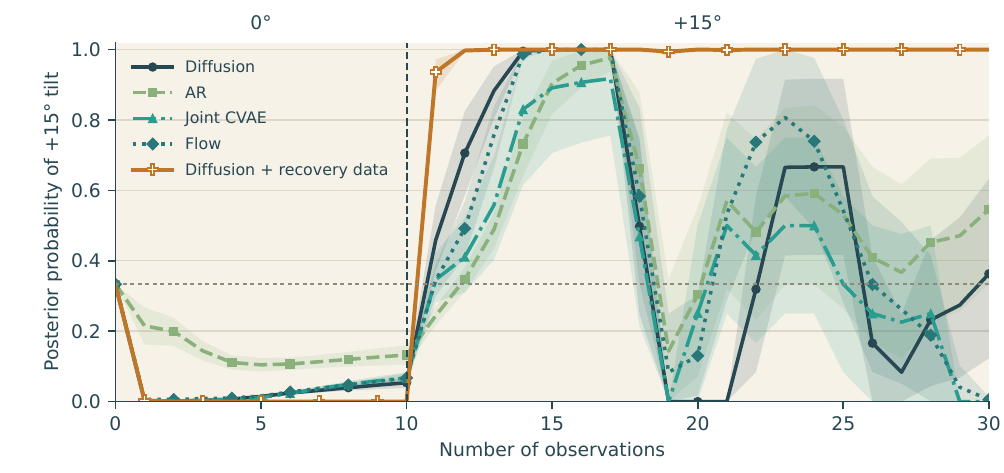}
    \caption{E2 replay: mean \(p(+15^{\circ})\) over 12 trials; observation 11 first reflects the switch. Shading: pointwise 95\% bootstrap intervals; dotted line: uniform prior.}
    \label{fig:e2_belief_tracking}
\end{figure}

\smallskip
\noindent\textbf{E2: Belief adaptation and recovery experience.}
We evaluate belief inference by replaying twelve trials in which lateral tilt switches from \(0^{\circ}\) to \(+15^{\circ}\) after ten actions. The four pretrained models share recorded actions and observations, with matched control trials at normal lateral tilt. An additional arm evaluates the same diffusion model after offline fine-tuning on tilt-change recovery demonstrations. Every arm uses replay: models update their beliefs but do not select or alter the recorded actions. Only the recovery variant receives additional training; all model weights remain fixed during evaluation. Starting from a uniform belief, each model predicts observations under three tilt hypotheses and applies
\begin{equation}
b_{t+1}(z')\propto
\ell_\theta(o_{t+1}\mid h_t,a_t,z')
\sum_z M_{z',z}b_t(z),
\label{eq:e2_filter}
\end{equation}
where \(z'\) governs the observed transition. The context-transition matrix \(M\) is fixed from development data and allows changes between hypotheses. The filter receives neither a switch indicator nor its timing, and uses no posterior reset or test-time refitting.

\begin{table}[!b]
    \caption{E2 belief tracking (12 trials per arm). Delay: post-switch observations to \(p(+15^{\circ})\geq0.9\); correctness uses the most likely tilt.}
    \label{tab:e2_tracking}
    \centering
    \small
    \setlength{\tabcolsep}{5pt}
    \begin{tabular}{lccc}
        \toprule
        Model & \shortstack{Identification delay\\median [min,max]} & \shortstack{Identified\\by six} & \shortstack{Correct at\\observation 30} \\
        \midrule
        Diffusion & 3 [2,4] & 12/12 & 5/12 \\
        Autoregressive & 5 [5,8] & 10/12 & 7/12 \\
        CVAE & 4 [3,10] & 10/12 & 0/12 \\
        Flow matching & 4 [3,4] & 12/12 & 0/12 \\
        \midrule
        \textbf{Diffusion + recovery data} & \textbf{1 [1,2]} & \textbf{12/12} & \textbf{12/12} \\
        \bottomrule
    \end{tabular}
\end{table}

The pretrained models correctly classify matched trials at normal lateral tilt over observations 10--30. After the switch, diffusion's belief adapts fastest among these models: median delay to 90\% confidence is three observations, versus four or five for the other families (Table~\ref{tab:e2_tracking}). Diffusion and flow identify the new tilt in all twelve trials within six observations. Both tilt values occur in the original training data, but within-trajectory switches do not. The rapid response therefore demonstrates adaptation of the hidden-state belief using a fixed learned model, even without demonstrations of the change itself. At observation 30, diffusion remains correct in 5/12 trials, ahead of CVAE and flow (0/12), although autoregression retains more correct classifications (7/12).

The later belief fluctuations in Fig.~\ref{fig:e2_belief_tracking} accompany the ball slipping away from the pusher in the replayed trajectories, while actions planned for normal lateral tilt continue. The original demonstrations cover fixed conditions but no recovery after sudden tilt changes. Once contact is lost, the resulting motion is unfamiliar to these models, and their predictions become less reliable as observation evidence. This explains why recognizing a familiar tilt value can be easier than maintaining an accurate belief through the motion that follows its sudden onset: training coverage must include the resulting states and transitions, not only the possible tilt values.

With recovery fine-tuning, diffusion rapidly identifies the new tilt from replayed observations and sustains accurate tracking. Median identification delay falls from three observations to one, final correctness rises from 5/12 to 12/12, and the mean posterior remains near one after identification. These results show the value of recovery demonstrations for learning observation evidence that remains useful after a dynamics change. Together with diffusion's strong action proposals and accurate controlled predictions in E1, its rapid belief response and the improvement with recovery data make it the best-supported candidate for GenAIF's shared world-model role among the tested implementations. E2 establishes the inference capability; evaluating how those beliefs guide adaptive actions is the next step.

\section{Discussion}
\label{sec:discussion}

Our vision for GenAIF is to learn a system's behaviour from rich trajectory demonstrations and use that understanding to connect inference with action. This paper takes a first step toward that vision: it derives the required model capabilities and benchmarks candidate families against them.

The central conclusion is that one shared trajectory model can provide a goal-conditioned policy distribution and an approximate state-to-observation likelihood mapping, supporting action proposals, controlled prediction, and belief revision. These capabilities must be evaluated together: plausible trajectories and accurate forecasts alone do not establish effective control. Diffusion offers more reliable proposals in the tested setting, while CVAE provides comparable short-horizon prediction at substantially lower query cost. Their suitability for GenAIF depends on the balance between proposal quality, prediction, evidence reliability, and computation. Accurate conditioning is fundamental to using either model effectively. The earlier proposal-only experiments in Appendix~\ref{app:legacy_proposals} provide complementary evidence for the action-search component.

E2 shows that recovery demonstrations strengthen belief inference after a dynamics change. The next step is to close the loop: use the inferred condition and \(G\) to select actions, then evaluate control adaptation and the value of information-seeking behaviour. We aim to implement GenAIF in complex industrial systems whose interacting dynamics make detailed models impractical to construct or maintain. Rich trajectory demonstrations would provide the main learning signal, supported by limited structural knowledge and a graph linking subsystem models. A feature-learning loop would extract hidden-state representations and operating conditions from these datasets, progressively reducing the need to specify the form and range of hidden dynamics in advance.

\clearpage
\begingroup
\setlength{\emergencystretch}{1em}
\appendix
\setcounter{figure}{0}
\setcounter{table}{0}
\renewcommand{\thefigure}{A\arabic{figure}}
\renewcommand{\thetable}{A\arabic{table}}
\renewcommand{\theHsection}{appendix.\arabic{section}}
\renewcommand{\theHfigure}{appendix.\arabic{figure}}
\renewcommand{\theHtable}{appendix.\arabic{table}}
\setlength{\textfloatsep}{10pt plus 1pt minus 2pt}
\setlength{\floatsep}{8pt plus 1pt minus 2pt}

\section{Proposal Models with Simulator-Based Planning}
\label{app:legacy_proposals}
\markboth{Proposal Models with Simulator-Based Planning}{Proposal Models with Simulator-Based Planning}

The main paper evaluates learned models as world models. This appendix instead tests proposal distributions with MuJoCo transitions, measuring candidate quality and finite-budget control when an accurate simulator is available. It provides practical guidance for model-based planners, but does not demonstrate learned dynamics or the complete GenAIF loop.

All controllers use simulator rollouts, common preference and safety scoring, an eight-step horizon, and receding-horizon execution. The learned proposals come from a separate archive of successful RL-specialist demonstrations; their data and checkpoints differ from E1 and E2. We compare them with the cross-entropy method (CEM) \cite{Rubinstein1999CEM,Botev2013CEM}. Six scenes and three sampling seeds give eighteen episodes per family and budget. ``Proposal model'' pools five separately evaluated controllers with equal family weight: BC-MDN \cite{Bishop1994MDN}, CVAE, diffusion, flow, and autoregressive. Gated controllers mix learned and fallback candidates.

\smallskip
\noindent\textbf{Learned proposals improve control with limited search.}
We vary the candidate budget while keeping rollout and scoring shared. At just eight candidates, learned proposals succeed in most trials, while CEM fails throughout (Fig.~\ref{fig:legacy_scaling}A). Gating raises reliability further and solves nearly every trial at the largest common budget. The improvement comes from supplying useful alternatives to the planner: a good transition model alone cannot compensate for a poor candidate set.

More extensive CEM search closes much of the success gap at greater planning cost (Fig.~\ref{fig:legacy_scaling}B). Because this configuration also doubles refinement iterations and changes the elite rule, it represents a stronger search rather than a candidate-count change alone. Learned proposals use small search budgets effectively; extensive simulator search remains a viable alternative.

\smallskip
\noindent\textbf{Diffusion favours candidate quality; CVAE favours speed.}
To separate proposal quality from candidate count, we compare all five families at \(K=128\) (Table~\ref{tab:legacy_families}). Diffusion has the highest fraction of feasible candidates and succeeds in every trial. CVAE nearly matches its success with much cheaper generation. The gap in total planning time is much smaller because simulator rollouts and scoring dominate the fast generators. Model selection should therefore use complete planning cost as well as proposal latency.

\begin{table}[!b]
    \caption{Proposal-family comparison at \(K=128\). Feasibility averages the fraction of feasible candidates per rollout; times are per decision.}
    \label{tab:legacy_families}
    \centering
    \small
    \setlength{\tabcolsep}{6pt}
    \begin{tabular}{lrrrr}
        \toprule
        Family & Success & Feasible (\%) & Proposal (ms) & Plan (ms) \\
        \midrule
        BC-MDN & 11/18 & 82.4 & 0.75 & 346 \\
        CVAE & 17/18 & 92.1 & 0.39 & 352 \\
        Diffusion & 18/18 & 100.0 & 50.05 & 411 \\
        Flow & 14/18 & 83.9 & 3.86 & 354 \\
        Autoregressive & 14/18 & 82.1 & 14.03 & 360 \\
        \bottomrule
    \end{tabular}
\end{table}

\begin{figure}[!b]
    \centering
    \includegraphics[width=\linewidth]{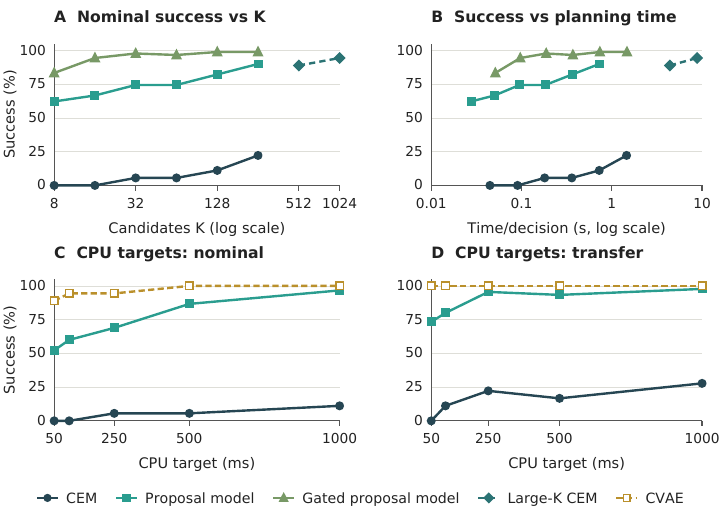}
    \caption{Proposal-only control with MuJoCo transitions. Success versus (A) candidate budget, (B) measured planning time, and (C,D) requested CPU targets for nominal and new start--goal scenes. Diffusion exceeds the two smallest CPU targets, so pooled comparisons there are not at equal measured cost.}
    \label{fig:legacy_scaling}
\end{figure}

\smallskip
\noindent\textbf{The advantage survives comparable computation and new layouts.}
We calibrate candidate counts to CPU planning targets and repeat the evaluation on six new start--goal scenes with the same three sampling seeds. On one Torch thread, we select the largest tested \(K\) whose 90th-percentile planning time fits the target, or the smallest \(K\) if none fits; offline training is excluded.

At the 50\,ms target, CVAE and CEM have nearly identical measured planning cost (Table~\ref{tab:legacy_cpu}), yet CVAE succeeds in most nominal trials and every transfer trial while CEM solves neither. The proposal advantage therefore persists under comparable computation and new layouts; CVAE also leads the wider low-latency sweep (Fig.~\ref{fig:legacy_scaling}C,D).

\begin{table}[!t]
    \caption{CEM and CVAE at a 50\,ms CPU target. Transfer uses new start--goal scenes; planning time includes proposal generation, rollout, and scoring.}
    \label{tab:legacy_cpu}
    \centering
    \small
    \setlength{\tabcolsep}{5pt}
    \begin{tabular}{lrrrrr}
        \toprule
        & & \multicolumn{2}{c}{Success /18} & \multicolumn{2}{c}{Plan time (ms)} \\
        \cmidrule(lr){3-4}\cmidrule(lr){5-6}
        Method & \(K\) & Nominal & Transfer & Nominal & Transfer \\
        \midrule
        CEM & 8 & 0 & 0 & 44.5 & 44.4 \\
        CVAE & 16 & 16 & 18 & 43.8 & 43.5 \\
        \bottomrule
    \end{tabular}
\end{table}

The CVAE--CEM result uses measured timings; all findings use fixed checkpoints and small scene sets.
\label{app:legacy_end}

\endgroup
\clearpage
\markboth{References}{References}

\end{document}